\documentclass[10pt,twocolumn,letterpaper]{article}

\usepackage{cvpr}              % To produce the CAMERA-READY version
\usepackage{multirow}
\usepackage[dvipsnames]{xcolor}

\definecolor{cvprblue}{rgb}{0.21,0.49,0.74}
\usepackage[pagebackref,breaklinks,colorlinks,citecolor=cvprblue]{hyperref}

\def\paperID{13} % *** Enter the Paper ID here
\def\confName{CVPR}
\def\confYear{2024}

\title{Analyzing Participants' Engagement during Online Meetings Using Unsupervised Remote Photoplethysmography with Behavioral Features}

\author{Alexander Vedernikov\textsuperscript{1}, Zhaodong Sun\textsuperscript{1}, Virpi-Liisa Kykyri\textsuperscript{2}, \\ Mikko Pohjola\textsuperscript{2}, Miriam Nokia\textsuperscript{2}, and Xiaobai Li\textsuperscript{3,1,}\thanks{Corresponding author.}\vspace{0.1cm} \\
\textsuperscript{1}{Center for Machine Vision and Signal Analysis, University of Oulu, Finland}\\
\textsuperscript{2}{Department of Psychology, University of Jyväskylä, Finland}\\
\textsuperscript{3}{State Key Laboratory of Blockchain and Data Security, Zhejiang University, China}\\ 
{\tt\small \{aleksandr.vedernikov, zhaodong.sun\}@oulu.fi}, {\tt\small xiaobai.li@zju.edu.cn}\\
{\tt\small \{virpi-liisa.kykyri, mikko.j.pohjola, miriam.nokia\}@jyu.fi}
}

\begin{document}
\maketitle
\begin{abstract}
Engagement measurement finds application in healthcare, education, services. The use of physiological and behavioral features is viable, but the impracticality of traditional physiological measurement arises due to the need for contact sensors. We demonstrate the feasibility of unsupervised remote photoplethysmography (rPPG) as an alternative for contact sensors in deriving heart rate variability (HRV) features, then fusing these with behavioral features to measure engagement in online group meetings. Firstly, a unique \textit{Engagement Dataset} of online interactions among social workers is collected with granular engagement labels, offering insight into virtual meeting dynamics. Secondly, a pre-trained rPPG model is customized to reconstruct rPPG signals from video meetings in an unsupervised manner, enabling the calculation of HRV features. Thirdly, the feasibility of estimating engagement from HRV features using short observation windows, with a notable enhancement when using longer observation windows of two to four minutes, is demonstrated. Fourthly, the effectiveness of behavioral cues is evaluated when fused with physiological data, which further enhances engagement estimation performance. An accuracy of 94\% is achieved when only HRV features are used, eliminating the need for contact sensors or ground truth signals; use of behavioral cues raises the accuracy to 96\%. Facial analysis offers precise engagement measurement, beneficial for future applications.
\end{abstract}
    
\section{Introduction}
\label{sec:intro}

The rapid transition to online communication underlines the importance of engagement analysis in virtual meetings. With the absence of physical cues and direct interaction between individuals, assessing engagement becomes more challenging. Nevertheless, estimating engagement during these interactions offers key insights into participant behavior, group dynamics, and individual input, helping meeting organizers promote effective collaboration.

\begin{figure}[t]
  \centering
  \includegraphics[width=0.65\linewidth]{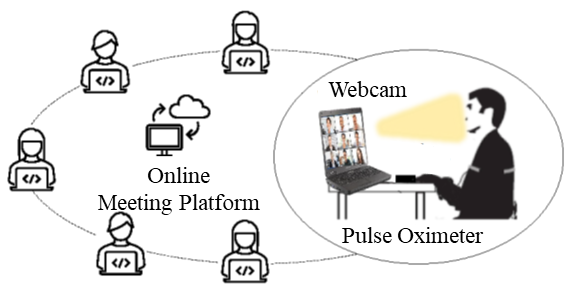}
  \caption{A large \textit{Engagement Dataset} of realistic online meetings. Facial videos and cPPG were recorded.}
  \label{fig:dataset}
\end{figure}

Engagement analysis in virtual meetings often relies on facial and body language recognition \cite{Chen2023smg,kamath2016crowdsourced}, although they can't gauge direct physiological responses like heart rate variability (HRV), which is difficult for individuals to fake \cite{sun2022estimating}. Electrocardiography (ECG) would be ideal but is limited by its need for direct contact. In contrast, remote photoplethysmography (rPPG), a computer vision-based technique, assesses cardiac activity through facial color changes.
This makes it suitable for online engagement estimation as it obviates the need for direct contact and specialized equipment \cite{Sabour2023ubfc}. Using unsupervised deep learning for rPPG signal extraction eliminates the need for ground truth signals, labeled datasets, and expert annotations, enhancing flexibility in engagement estimation.

This article pioneers the application of unsupervised rPPG measurement technology in estimating engagement during online meetings. It also proposes an enhancement in engagement estimation by integrating behavioral cues such as facial expression and motion. This study discusses potential impediments affecting the performance of rPPG in engagement analysis. A significant influence on engagement estimation performance arises from the size of the observation window employed for calculating HRV features from the obtained rPPG signal. Moreover, a \textit{Engagement Dataset} of online group meetings (captured in real-world scenarios) among social workers is collected (\cref{fig:dataset}), the first to explore engagement levels in group interactions, especially with many participants. It contains 1.5-hour videos for in-depth engagement analysis, unveiling evolving interactions and gradual shifts in group dynamics, and incorporates heart rate (HR) data from contact PPG (cPPG), a feature absent in public datasets. By using a continuous engagement range of -10 to +10, the \textit{Engagement Dataset} captures nuances that tend to be neglected in other datasets \cite{gupta2016daisee,kaur2018prediction,lee2022predicting}. The analysis reveals that participant engagement is linked to work effectiveness and mitigates job stress, underlining the study's practical value.

\section{Related work}
\label{sec:formatting}

Automated engagement analysis research, initiated by Whitehill \etal~\cite{Whitehill201486} in 2014, showcased machine learning's capacity to estimate engagement with human-like precision.
Subsequent studies explored a wide range of applications \cite{kumar2024nontypical}, including education \cite{savchenko2022classifying}, social media \cite{eslami2022understanding}, news \cite{hoque2022analyzing}, human-robot \cite{salam2022automatic} and human-human interaction \cite{celiktutan2017multimodal}, consumer engagement \cite{eslami2022understanding}, healthcare \cite{steinert2021audio}, games \cite{chen2019faceengage}, and film viewing analysis \cite{benini2019influence}. These studies explored behavioral features like facial expressions \cite{gordon2016affective}, eye gaze \cite{choi2022immersion}, and body gestures \cite{khenkar2022engagement}. There were studies that utilized physiological features, \eg, \cite{shaw20221d} used features extracted from electroencephalogram (EEG) signals. In addition, studies also analyzed modalities reliant solely on text \cite{atapattu2019detecting}, reaction time \cite{ober2021detecting}, and response accuracy \cite{ober2021detecting}. Rather than using a single modality, scientists have investigated multi-modality fusion methods merging facial expression-related features with speech/audio \cite{Huang2016590}, head and body pose/gestures/motions \cite{psaltis2017multimodal}, physiological signals (such as electroencephalographic - EEG activity \cite{arapakis2017interest}, thermal signals \cite{filippini2021facilitating}, and electrodermal activity - EDA \cite{dubovi2022cognitive}), game events \cite{psaltis2017multimodal}, mouse behavior \cite{zhang2020data}, and contextual information \cite{aslan2019investigating}. Past research on engagement estimation is robust, but fails to leverage the improvements rPPG data can provide. Notably, rPPG technology has been applied in affective computing, showing promise in assessing emotional states such as depression \cite{Casado2023depression}, stress \cite{Sabour2023ubfc}, and embarrassment \cite{wu2023recognizing}, indicating unexploited potential for engagement analysis.

Over the past decade, only one study in 2016 by Monkaresi \etal~\cite{monkaresi2016automated} explored HR for engagement estimation, but it faced major constraints. Firstly, they relied on contact ECG sensors for HR, impractical in real use. Our approach, however, uses a non-intrusive method, eliminating contact sensors. Secondly, the video methods of that time \cite{hsu2014learning} confined their research to lab data, suffering in real HR detection scenarios. Conversely, our method is tested and effective in real-life conditions. Moreover, using only seven basic HR statistical features, they failed to fully exploit HR signals' potential, leading to a high clinical error rate, as the authors acknowledged. Their experiments showed that HR signals were less effective than facial expressions in estimating engagement, likely due to unreliable cardiac information. HRV features, strongly linked to mental states \cite{liang2009changes}, offer potential as efficient engagement indicators demanding more research, driving this work's innovative approach to physiological signals.

As engagement estimation methods advanced, multiple datasets were created. Real-world e-learning engagement was first examined by the \textit{DAiSEE} dataset \cite{gupta2016daisee} in 2016. The horizons of student engagement analysis in educational games were broadened by the \textit{Multimodal Affective State Recognition Dataset} \cite{psaltis2017multimodal}. Using the \textit{MHHRI} dataset \cite{celiktutan2017multimodal}, engagement in dyadic human and triadic human-human-robot contexts was explored. In 2017, \textit{UE-HRI} dataset \cite{ben2017ue} further expanded the scope of human-robot interaction and centered around interactions with the robot Pepper. In the domain of e-learning, the \textit{EngageWild} dataset \cite{kaur2018prediction} and \textit{VRESEE} datasets \cite{selim2022students} delved deep into students' engagement patterns. YouTube gaming videos and facial engagement modalities were uniquely integrated in the \textit{FaceEngage} dataset \cite{chen2019faceengage}. A predictive direction in the field was signified by the \textit{PAFE} dataset \cite{lee2022predicting} in 2022 and the EngageNet dataset \cite{singh2023engagenet} in 2023, both utilizing different contexts. Primarily, all mentioned datasets offer only visual modality for engagement analysis, with just two exceptions: the \textit{MHHRI} dataset \cite{celiktutan2017multimodal} includes data from audio, video, depth, electrodermal activity (EDA), temperature, and 3-axis wrist acceleration, while the \textit{UE-HRI} dataset \cite{ben2017ue} delivers information from a microphone array, cameras, depth sensors, sonars, lasers, and user feedback captured through the robot's touchscreen. The lack of cPPG data in public resources for rPPG engagement methods underlines the necessity for new datasets and unsupervised approaches.

This paper introduces the application of unsupervised deep learning for calculating rPPG signals, an unexplored modality in previous automated engagement analyses, emphasizing its practical, non-contact, and non-intrusive nature. Furthermore, various behavioral feature sets are also evaluated and fused with physiological features to further boost the performance. The work is established on a novel self-collected \textit{Engagement Dataset}, capturing real-world online video meetings of social workers with consultant therapists for the purpose of reducing work-related stress. The \textit{Engagement Dataset} construction, the proposed method framework, and experimental results are explained in the following sections. 
\section{Dataset construction}
\label{sec:dataset_construction}

\begin{figure*}[t]
  \centering
   \includegraphics[width=0.80\linewidth]{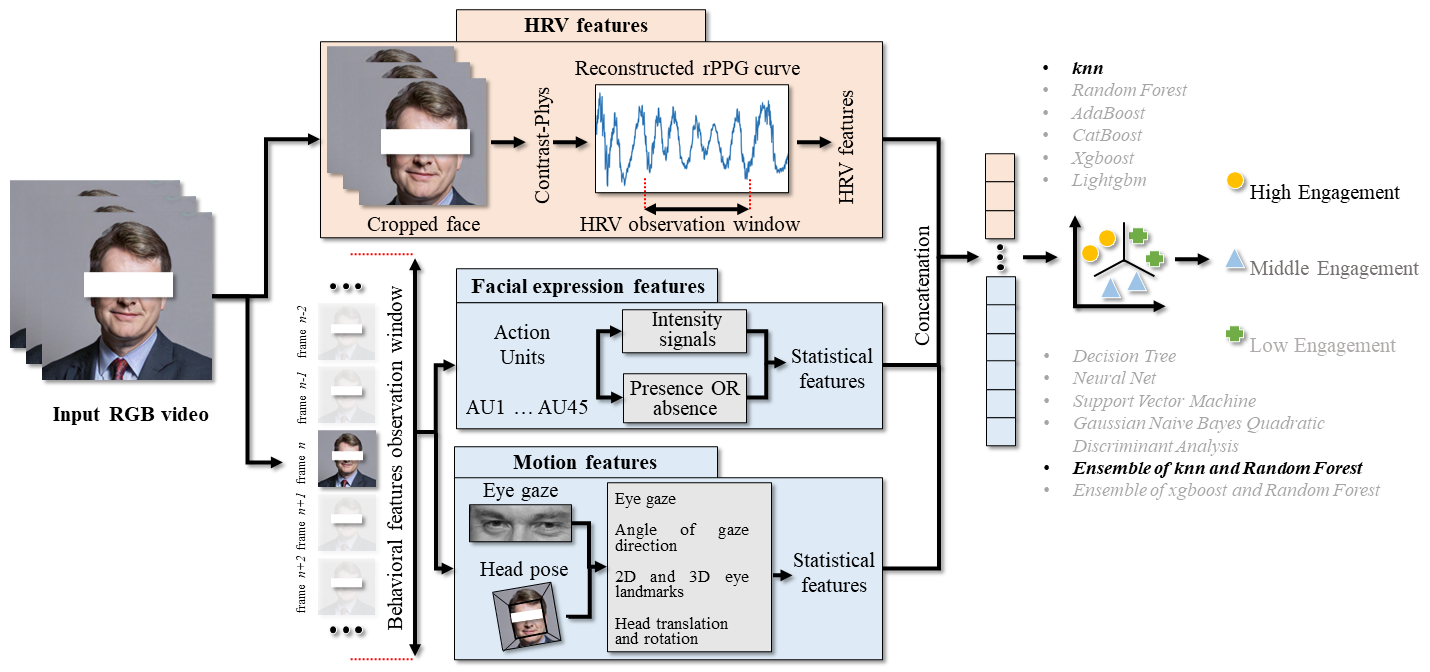}
   \caption{The diagram depicts engagement estimation based on heart rate variability (HRV) and behavioral features. HRV features are calculated from reconstructed rPPG signals, and behavioral features are computed using the OpenFace library \cite{baltrusaitis2018openface}. Feature-level multimodality fusion is employed, followed by a classifier.}
   \label{fig:modelperf}
\end{figure*}

\subsection{Data collection}
The \textit{Engagement Dataset}, which contains facial videos, cPPG data, and engagement annotations, was collected (\cref{tab:datasettable}). The \textit{Engagement Dataset} comprises recordings from group online video meetings. Serving as a reflection of participants' everyday work, each online video meeting was organized using the Zoom platform and involved the participation of seven to nine individuals with one consultant. These participants were social services employees working with individuals facing mental health challenges and substance abuse issues. Prior to each session, participants were given detailed instructions on how to set up their recording environment. To stream and record participants' facial videos during online meetings, webcams were installed, and OBS Studio was used. Throughout the sessions, a research assistant provided online guidance using Zoom Breakout Rooms. Participants were encouraged to freely express themselves and move, as long as their faces remained visible in the videos. The cPPG signals were captured using the Beurer 80 pulse oximeter. Following each online video meeting, the facial videos and cPPG signals were synchronized based on their timestamps.

\begin{table}[b]
  \centering
    {\fontsize{9}{8.5}\selectfont
  \begin{tabular}{lc}
    \toprule
    \textbf{Engagement labels} & [-10, 10], 2 labels/second \\
    \textbf{Engagement classes} & 3 (Low / Middle / High) \\
    \textbf{Participants (male/female)} & 25 (3/22) \\
    \textbf{Consultants (male/female)} & 2 (1/1) \\
    \textbf{Video recordings} & 109 \\
    \textbf{cPPG data} & 106 \\    
    \textbf{Average duration} & 1 hr 23 min 58 sec \\
    \textbf{Total length} & 153 hours \\
    \bottomrule
  \end{tabular}
  \caption{\textit{Engagement Dataset} statistics and properties.}
  \label{tab:datasettable}}
\end{table}

\subsection{Data annotation}
Psychology students worked as research assistants to perform engagement annotation based on the observed behavior of the target subject from face video. DARMA, a continuous measurement system, was used for engagement annotation. It synchronizes media playback and continuous recording of observational measurement conducted with a computer joystick at a sampling rate of 20Hz. The DARMA software provided a continuous coding time series consisting of two values per second \cite{leite2015comparing}, with 10 (“High Engagement") and -10 (“Low Engagement"). Research assistants underwent training to familiarize themselves with distinct engagement levels before starting the coding. Inter-rater reliability in control sessions was good (correlation coefficient over 0.8). 

The scale of -10 to +10 facilitated a detailed understanding of engagement, allowing for robust and nuanced annotations. This range was chosen over a binary or three-point scale because human emotional states and levels of engagement are continuous and nuanced. The chosen range captures these nuances by providing a more granular scale. It is broad enough to encompass the extreme ends of the engagement spectrum while being fine-grained enough to account for slight variations within these extremes. Additionally, this continuous scale can better account for individual differences in engagement levels, allowing for the representation of each subject’s unique engagement signature.

\subsection{Data statistics and properties}
The \textit{Engagement Dataset} comprises 24 recorded group online video meetings, each lasting approximately 1.5 hours. The \textit{Engagement Dataset} contains two modalities, namely facial video recordings and cPPG signals. Each online video meeting involved between seven to nine participants and one consultant. The resolution of 1920 × 1080 and the frame rate of 60 fps were recommended to optimize video quality, although variations were observed due to different cameras and recording environments. The \textit{Engagement Dataset} contains several unique properties as opposed to previous datasets studying engagement. 1) \textbf{Duration of data.} With 1.5-hour videos compared to shorter public clips, the \textit{Engagement Dataset} enables in-depth engagement analysis, uncovering evolving interactions, gradual shifts in interactions, and group dynamics. 2) \textbf{Group dynamics.} Unlike earlier datasets, the \textit{Engagement Dataset} involves more participants per session, amplifying group interaction complexities for better engagement analysis and enhancing group dynamic insights. 3) \textbf{Recorded cPPG signals.} The presence of cPPG signals in the \textit{Engagement Dataset}, not presented in other datasets, underscores its unique importance in advancing engagement estimation. 4) \textbf{Granular annotation.} The \textit{Engagement Dataset} uniquely adopts a -10 to +10 range of engagement, emphasizing the fluidity of human responses. Such a spectrum effectively captures engagement nuances often overlooked in other datasets. 5) \textbf{Real-world setting.} Rooted in real-world scenarios, the \textit{Engagement Dataset} showcases authentic online meetings of social service employees. Unlike other datasets in simulated and well-controlled environments, the \textit{Engagement Dataset} promises true engagement data pertinent to real-world situations. 
\section{Method}

\subsection{rPPG curves reconstruction from facial videos}
The proposed method is illustrated in \cref{fig:modelperf}. The original videos are pre-processed to extract frames and acquire facial landmarks using the OpenFace library \cite{baltrusaitis2018openface}. This tool effectively addresses issues related to head motion, providing precise tracking of facial landmarks. The detected facial landmarks are employed to determine regions of interest (ROI) on exposed skin areas. The faces are cropped and then resized to dimensions of 128 × 128, preparing them for input into the unsupervised Contrast-Phys model \cite{Sun2022contrast}, a computer vision-based technique, for the subsequent rPPG curves reconstruction. The Contrast-Phys approach is based on the principle where the utilization of a 3DCNN model allows the derivation of multiple rPPG signals from distinct spatiotemporal locations within each video. Two videos, randomly selected from the \textit{Engagement Dataset}, constitute the input of Contrast-Phys. One video yields spatiotemporal rPPG (ST-rPPG) block $P$, rPPG samples $[p_1, \dots, p_N]$, and associated power spectrum densities (PSDs) $[f_1, \dots, f_N]$. The other video provides ST-rPPG block $P^{\prime}$, rPPG samples $[p^{\prime}_1, \dots, p^{\prime}_N]$, and corresponding PSDs $[f^{\prime}_1, \dots, f^{\prime}_N]$, following the same procedure. The contrastive loss pulls together PSDs from the same video while pushing apart PSDs from distinct ones. Contrast-Phys implies that using rPPG spatiotemporal similarity, the PSDs from the same ST-rPPG block should resemble each other as follows $\text{PSD}\big\{P(t_1 \to t_1+\Delta t, h_1, w_1)\big\} \approx \text{PSD}\big\{P(t_2 \to t_2+\Delta t, h_2, w_2)\big\} \implies f_i \approx f_j, i \neq j$ and $\text{PSD}\big\{P^{\prime}(t_1 \to t_1+\Delta t, h_1, w_1)\big\} \approx \text{PSD}\big\{P^{\prime}(t_2 \to t_2+\Delta t, h_2, w_2)\big\} \implies f^{\prime}_i \approx f^{\prime}_j, i \neq j$. Subsequently, to bring together PSDs (positive pairs) from the same video, the mean squared error is suggested to be utilized as the loss function. When normalized based on the total count of positive pairs, the positive loss term is: \vspace{-0.22cm}

{
\fontsize{9pt}{6pt}\selectfont
\begin{equation}
L_p =  \sum_{i=1}^{N} \sum_{\substack{j=1 \\j \neq i}}^{N} \big(\parallel f_i - f_j \parallel^2 + \parallel f^{\prime}_i - f^{\prime}_j \parallel^2 \big) / \big(2N(N-1)\big)
\end{equation}}

On the other hand, the cross-video rPPG dissimilarity suggests that the PSDs resulting from spatiotemporal sampling of two separate ST-rPPG blocks will be distinct. This attribute for the two input videos is described as $\text{PSD}\big\{P(t_1 \to t_1+\Delta t, h_1, w_1)\big\}
\neq \text{PSD}\big\{P^{\prime}(t_2 \to t_2+\Delta t, h_2, w_2)\big\} \implies f_i \neq f^{\prime}_j$. Next, the task of distancing PSDs (negative pairs) from two different videos is achieved when the negative mean squared error is used as the loss function. Then, the overall quantity of negative pairs is employed to normalize the negative loss term: 

{
\fontsize{9pt}{6pt}\selectfont
\begin{equation}
L_n = - \sum_{i=1}^{N} \sum_{j=1}^{N} \parallel f_i - f^{\prime}_j \parallel^2  / N^2
\end{equation}}

Finally, the overall loss function combines both positive and negative loss terms: $L = L_p + L_n$.

The model is pre-trained on the Oulu Bio-Face database \cite{li2018obf}. This approach enables the reconstruction of rPPG signals in any recorded facial video, eliminating the need for ground truth data in the future.

%-------------------------------------------------------------------------
\subsection{Definition of observation window for HRV and behavioral features}

The -10 to +10 engagement scale is divided into three classes - Low, Medium, and High engagement - through a process designed to simplify the complexity of engagement analysis while retaining a significant level of detail. This triadic classification approach provides a balance between the -10 to +10 scale's granularity and a binary scale's simplicity, serving as a practical and efficient method for multi-classifying engagement levels. In High engagement (Score: 10), the participant is either speaking, attempting to take a turn, speaking over someone else, or, as a listener, actively showing engagement through minimal vocalizations (such as 'mmm'), nods, and/or facial expressions. In Medium engagement (Score: 0), the participant either gazes at the speaker and appears to be listening, or gazes away while still seeming attentive. Finally, in the Low engagement (Score: -10), 
the participant gazes away or has their eyes closed, appearing not to follow the conversation, possibly engaging in side activities like turning away or opening emails. The process of converting continuous engagement labels into three classes and definition of observation windows for HRV and behavioral features (BF) is detailed below.

\begin{figure}[b]
  \centering
   \includegraphics[width=0.75\linewidth]{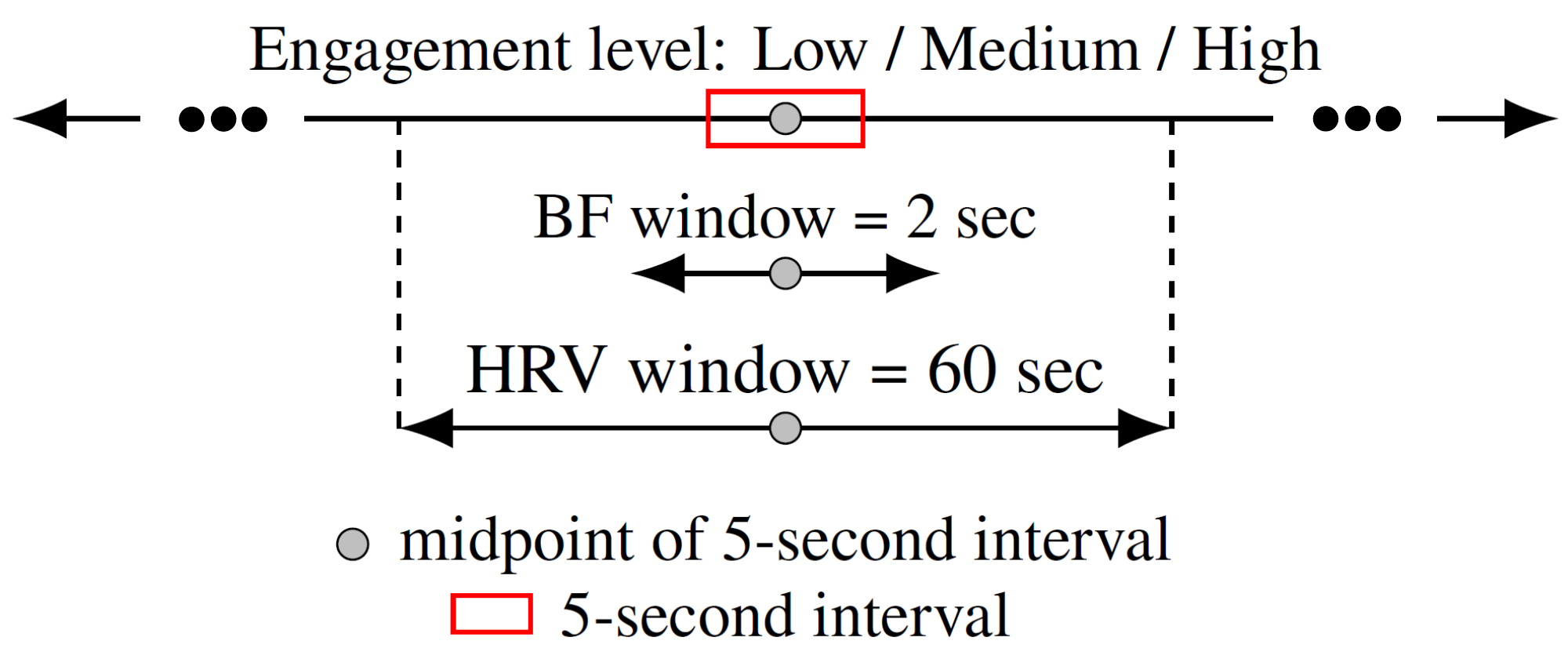}
   \caption{Observation windows concept for extracting BF and HRV features in engagement analysis.}
   \label{fig:window}
\end{figure}

A 5-seconds time period was chosen for detecting intervals of stable engagement. Reliability risks arise from short, context-lacking clips. Assessing longer clips is more complex due to mixed engagement levels. Therefore, video recordings were partitioned into 5-seconds intervals, and the standard deviations of engagement were computed for each interval. The median value of calculated standard deviations across the collected \textit{Engagement Dataset} was subsequently estimated. This median served as a threshold for filtering out 5-seconds intervals with higher engagement standard deviations. This process excluded unstable intervals with significant engagement fluctuations. The engagement values obtained from the filtered intervals were classified as follows: -10 to 0 as 'Low engagement', 0 to 5 as 'Medium engagement', and 5 to 10 as 'High engagement'. Example of engagement data of a specific participant in a particular video meeting after filtering process is shown in \cref{fig:eng_individual} (see supplementary material). After the filtering process, the distribution of the samples was as follows: 4001 samples were categorized as 'Low Engagement', 23069 as 'Medium Engagement', and 8320 as 'High Engagement'. Next, the midpoint of each 5-seconds filtered interval becomes the center of an HRV and behavioral observation window (\cref{fig:window}). Subsequently, HRV features are computed with a 60-seconds observation window that requires a 30-seconds step in both directions, while BF is derived using a 2-seconds observation window following similar logic.

%-------------------------------------------------------------------------
\subsection{Feature extraction}
\textbf{HRV features.} Reconstructed rPPG signals were processed using the Neurokit2 library \cite{makowski2021neurokit} to identify systolic peaks. This allowed for inter-beat intervals (IBIs) calculation, yielding HRV features. These included three Poincaré plot features, 16 time-domain features, and five frequency-domain features, making a total of 24 HRV features. The impact of HRV observation window size on engagement estimation performance was investigated by extracting HRV features from the reconstructed rPPG signals using various observation window sizes (60, 90, 120, 150, 180, 210, and 240 seconds).

\noindent\textbf{Facial expression features.}
Action Units (AUs) were used for the computation of facial expression features \cite{tian2001recognizing}. 17 unique AUs were detected and tracked using the OpenFace library \cite{baltrusaitis2018openface} (see supplementary material \cref{tab:AUDescription}). 
AUs were represented in two ways: as intensity signals (on a scale from 0 to 5) and as binary classifications for their presence or absence. Hence, 34 features were obtained, providing a comprehensive insight into engagement-related facial movements. 

\noindent\textbf{Motion features.}
Four distinct sets of motion features were extracted using the OpenFace library \cite{baltrusaitis2018openface}. These included: six gaze tracking features illustrating the 3D coordinates of each eye's gaze direction; two features representing each eye's gaze direction; 280 features outlining 2D and 3D landmarks around each eye; and six features pertaining to the translation and rotation of the head in 3D space. Collectively, these 294 features provided a comprehensive examination of facial and eye movements associated with engagement. 

In the following, 'behavioral features' (BF) denotes motion and facial expressions. These BF were computed for each frame extracted from the collected video recordings. To consider the temporal dynamics of these features, a 2-seconds observation window was adopted, wherein the average value of each feature was computed. 

%-------------------------------------------------------------------------
\subsection{Engagement classification}
Three-class classification of engagement has been conducted in two stages. In the first stage, classification was done using only 24 HRV features with various classifiers. These included knn, Random Forest, AdaBoost, CatBoost, xgboost, lightgbm, Support Vector Machine (with 'poly', 'rbf', and 'sigmoid' kernels), Decision Tree, Gaussian Naive Bayes, Quadratic Discriminant Analysis, Neural Net, an ensemble of knn and Random Forest, and an ensemble of xgboost and Random Forest. Out of these, the knn and the ensemble of knn and Random Forest (knn+RF) proved to be the most effective. In the second stage, these two most effective classifiers were employed. A feature-level fusion technique was used to combine 24 HRV and 328 BF features into a single feature vector for a specific 5-seconds engagement interval, which was then fed into the classifier. 
\vspace{-0.1cm}
\section{Results}

%-------------------------------------------------------------------------
\subsection{Experimental protocol}
\label{subsec:exp_protocol}
The \textit{Engagement Dataset} was split into 80\% training and 20\% test sets. To ensure model robustness and generalization, a 5-fold subject dependent cross-validation was applied to the training data. The trained model predicts outcomes for the testing set. This protocol reliably measures model performance while maintaining test set independence \cite{scikitlearn2022}. Accuracy and the Receiver Operating Characteristic Area Under the Curve (ROC AUC) metrics were initially used to evaluate models. For the best performing models, the F1 score and confusion matrices were additionally calculated to enhance the comprehensiveness of the assessment.

%-------------------------------------------------------------------------
\subsection{HR measurement accuracy}
For the \textit{Engagement Dataset}, the HR from the reconstructed rPPG signals was compared to that of ground truth cPPG using mean absolute error (MAE) and root mean square error (RMSE) as evaluation metrics. An MAE of 5.15 bpm and an RMSE of 7.81 bpm were obtained, indicating promising results given the uncontrolled conditions of the video recordings. The SOTA performance \cite{Sabour2023ubfc} was marked by MAE values of 3.55, 5.99, and 9.26 bpm in controlled lab settings.

\begin{figure*}[ht]
  \centering
  \begin{subfigure}{0.40\linewidth}
    \includegraphics[width=\linewidth]{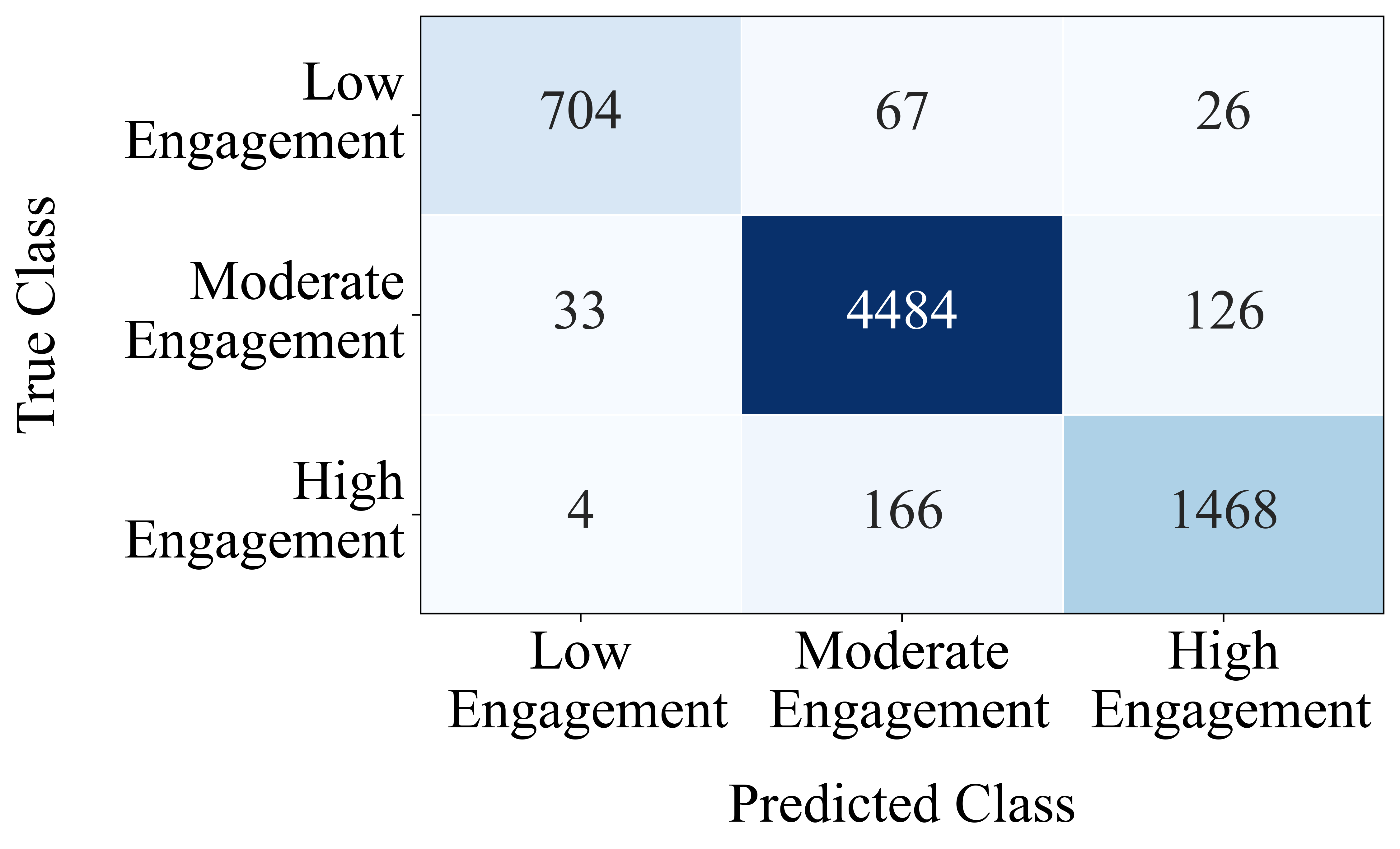} % Image file name inserted here
    \caption{Utilization of HRV features.}
    \label{fig:CM-a}
  \end{subfigure}
  \hfill
  \begin{subfigure}{0.40\linewidth}
    \includegraphics[width=\linewidth]{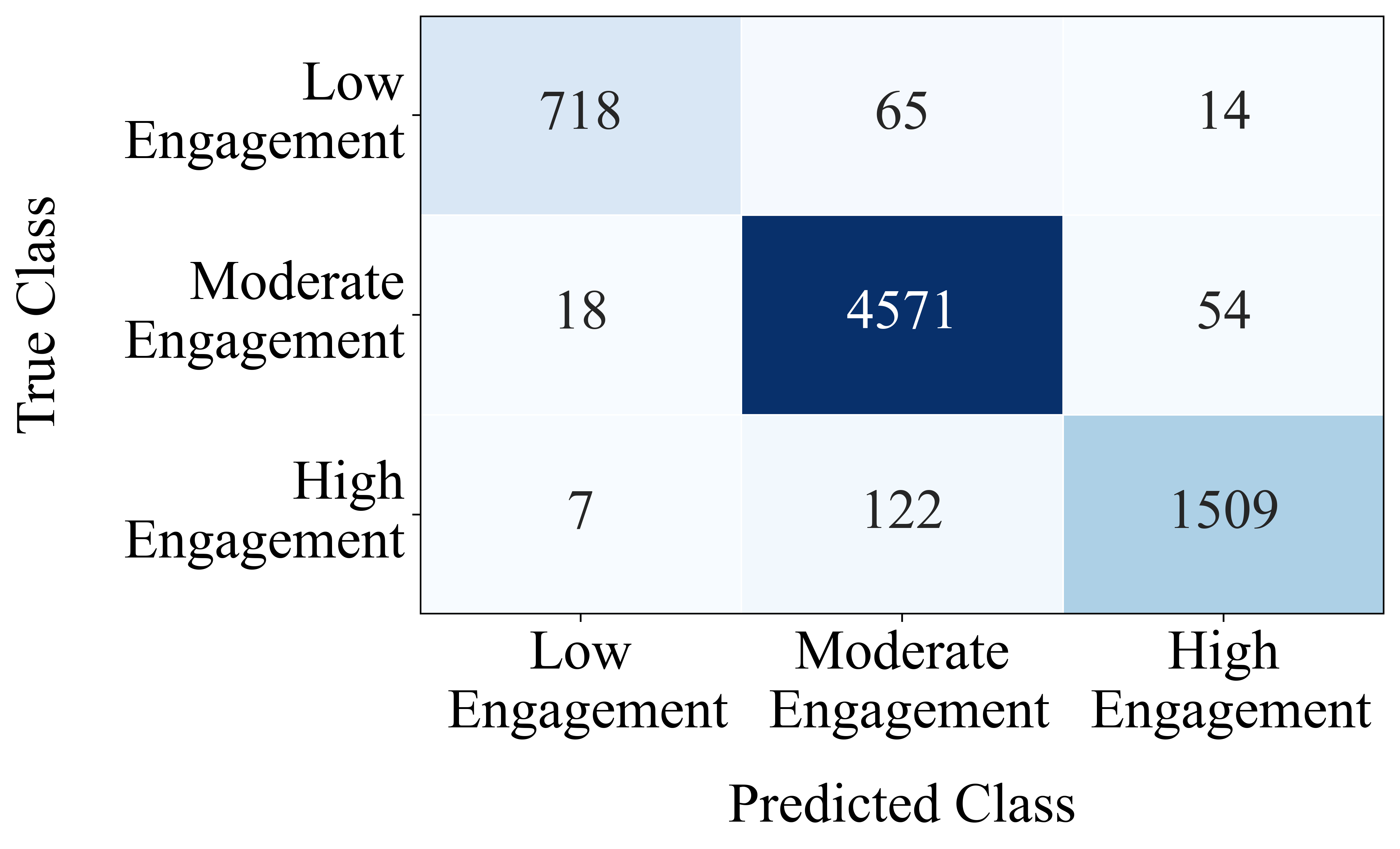} % Image file name inserted here
    \caption{Utilization of HRV and Behavioral features.}
    \label{fig:CM-b}
  \end{subfigure}
  \caption{Ensemble of knn and Random Forest model's confusion matrices for engagement estimation (HRV observation window of 240 seconds) based on (a) HRV features; (b) HRV and Behavioral features.}
  \label{fig:CM}
\end{figure*}
%-------------------------------------------------------------------------
\subsection{Short observation window HRV features for engagement estimation}
Utilizing a short observation window for HRV features calculation, the proposed approach's performance on the \textit{DAiSEE} \cite{gupta2016daisee} dataset (focused on students' engagement) and \textit{Engagement Dataset} is shown in \cref{tab:DAiSEE}. For the \textit{DAiSEE} dataset with 10-seconds snippets, the highest accuracy of 54.49\% was achieved by the Random Forest model. Meanwhile, a 49.40\% accuracy was yielded by the knn model on \textit{Engagement Dataset} using a 10-seconds observation window. The performance of the proposed method is constrained by the 10-seconds snippets of the \textit{DAiSEE} dataset. Reliable HRV features are not offered by such a short observation window, a fact also reflected in \textit{Engagement Dataset} performance using a 10-seconds observation window. Longer videos are required for robust HRV and engagement estimation. This study's video recordings allow the proposed method's capability to be fully realized using long HRV observation windows. In the following subsection, the influence of HRV observation window size is analyzed in detail.

\begin{table}[t]
  \centering
  {\fontsize{9}{9}\selectfont
  \begin{tabular}{l@{}c@{}}
    \toprule
    Method & Accuracy [\%] \\
    \midrule
    \multicolumn{2}{c}{\textit{DAiSEE} \cite{gupta2016daisee}} \\
    \midrule
    InceptionNet Video Level \cite{gupta2016daisee} & 46.40 \\
    InceptionNet Frame Level \cite{gupta2016daisee} & 47.10 \\  
    C3D Training \cite{gupta2016daisee} & 48.60 \\
    Inflated 3D ConvNet \cite{zhang2019annovel} & 52.35 \\
    % \underline{\textbf{Conformer}} & \underline{\textbf{52.35}} \\
    ResNet + TCN (weighted loss) \cite{abedi2021improving} & 53.70 \\
    \underline{\textbf{HRV + Random Forest}} & \underline{\textbf{54.49}} \\
    C3D FineTune \cite{gupta2016daisee} & 56.10 \\
    C3D LRCN \cite{gupta2016daisee} & 57.90 \\
    DFSTN \cite{liao2021deep} & 58.84 \\
    C3D + TCN \cite{abedi2021improving} & 59.97 \\
    ResNet + LSTM \cite{abedi2021improving} & 61.15 \\
    ResNet + TCN \cite{abedi2021improving} & 63.90 \\
    EfficientNet B7 + TCN \cite{selim2022students} & 64.67 \\
    EfficientNet B7 + Bi-LSTM \cite{selim2022students} & 66.39 \\
    Ordinal TCN \cite{abedi2021affect} & 67.40 \\
    EfficientNet B7 + LSTM \cite{selim2022students} & 67.48 \\
    \midrule
    \multicolumn{2}{c}{\textit{Engagement Dataset}} \\
    \midrule
    \underline{\textbf{HRV + knn}} & \underline{\textbf{49.40}} \\
    \bottomrule
  \end{tabular} 
  \caption{Proposed method's performance using short observation window HRV features for engagement estimation based on two datasets: (1) \textit{DAiSEE} \cite{gupta2016daisee}; (2) \textit{Engagement Dataset}.}
  \label{tab:DAiSEE}}
\end{table}

\begin{table}[b]
  \centering
      {\fontsize{9}{8.25}\selectfont

  \begin{tabular}{@{}ccccc@{}}
    \toprule
    HRV & \multicolumn{2}{c}{Accuracy [-]} & \multicolumn{2}{c}{ROC AUC [-]} \\
    window {[}sec{]} & knn & knn+RF & knn & knn+RF \\
    \midrule
    60    & 0.816 & 0.816 & 0.870 & 0.906 \\
    90    & 0.882 & 0.880 & 0.923 & 0.947 \\
    120   & 0.910 & 0.911 & 0.946 & 0.967 \\
    150   & 0.924 & 0.926 & 0.953 & 0.975 \\
    180   & 0.931 & 0.936 & 0.955 & 0.977 \\
    210   & 0.937 & 0.938 & 0.961 & 0.983 \\
    240   & 0.937 & \underline{\textbf{0.940}} & 0.960 & \underline{\textbf{0.983}} \\
    \bottomrule
  \end{tabular}
  \caption{Evaluation metrics of the knn model and ensemble of knn and Random Forest model for engagement estimation based on HRV features calculated at different values of HRV observation window.}
  \label{tab:modelperf}}
\end{table}

%-------------------------------------------------------------------------
\subsection{Effects of HRV observation window size on the performance of engagement estimation}

\begin{table*}[ht]
  \centering
      {\fontsize{9}{8.25}\selectfont
  \setlength\tabcolsep{5pt}  % adjust this value as needed
  \begin{tabular}{@{}ccccccccccc@{}}
    \toprule
    \begin{tabular}{@{}c@{}}HRV \\ window [sec]\end{tabular} 
    & HRV 
    & \begin{tabular}{@{}c@{}}HRV \\ + BF (1)\end{tabular}
    & \begin{tabular}{@{}c@{}}HRV \\ + BF (2)\end{tabular}
    & \begin{tabular}{@{}c@{}}HRV \\ + BF (3)\end{tabular}
    & \begin{tabular}{@{}c@{}}HRV \\ + BF (4)\end{tabular}
    & \begin{tabular}{@{}c@{}}HRV \\ + BF (5)\end{tabular}
    & \begin{tabular}{@{}c@{}}HRV \\ + BF (1+2)\end{tabular} 
    & \begin{tabular}{@{}c@{}}HRV \\ + BF (4+5)\end{tabular} 
    & \begin{tabular}{@{}c@{}}HRV \\ + BF (all)\end{tabular} \\ 
    
    \midrule
    60 & 0.816 & 0.864 & 0.855 & 0.861 & 0.896 & 0.919 & 0.856 & 0.928 & 0.930 \\

    90 & 0.880 & 0.903 & 0.902 & 0.870 & 0.936 & 0.931 & 0.891 & 0.942 & 0.929 \\

    120 & 0.911 & 0.929 & 0.929 & 0.881 & 0.952 & 0.944 & 0.919 & 0.952 & 0.934 \\

    150 & 0.926 & 0.934 & 0.936 & 0.884 & 0.953 & 0.944 & 0.927 & 0.954 & 0.940 \\

    180 & 0.936 & 0.937 & 0.944 & 0.892 & 0.954 & 0.947 & 0.932 & 0.955 & 0.938 \\

    210 & 0.938 & 0.939 & 0.944 & 0.896 & 0.955 & 0.951 & 0.933 & 0.958 & 0.941 \\

    240 & 0.940 & 0.943 & 0.949 & 0.899 & 0.956 & 0.956 & 0.934 & \underline{\textbf{0.960}} & 0.940 \\

    \bottomrule
  \end{tabular}
  \caption{Ensemble of knn and Random Forest model's accuracy for engagement estimation based on HRV and various sets of BF at different values of HRV observation window. (1) stands for gaze tracking, (2) for angle of gaze direction, (3) for 2D and 3D landmarks of specific points around each eye, (4) for translation and rotation of the head in 3D space, and (5) for facial Action Units.}
  \label{tab:big_comparison_table}}
\end{table*}

With the constraints of a 10-seconds observation window addressed in the preceding section, the effects of HRV observation windows of extended lengths are analyzed. The experiments revealed that knn and knn and Random Forest ensemble (knn+RF) were the most effective machine learning classifiers. Models' performance was assessed across different HRV observation window values (\cref{tab:modelperf}). As the HRV observation window size increased from 60 seconds to 240 seconds, significant improvements were observed in the accuracy and ROC AUC scores for both models. At an HRV observation window size of 60 seconds, the knn model achieved an accuracy of 0.816 and a ROC AUC score of 0.870, while the knn+RF model showed slightly better results with an accuracy of 0.816 and a ROC AUC score of 0.906. At the optimal HRV observation window size of 240 seconds, accuracy of the knn model reached 0.937 and its ROC AUC score peaked at 0.960, while the knn+RF model outperformed with an accuracy of 0.940 and a ROC AUC of 0.983. The HRV observation window size has a significant impact on the performance of engagement estimation models and larger observation window sizes yield better results. With longer observation windows, finer and more complex HRV patterns in the rPPG signals can be discerned, thus enabling more accurate and robust predictions based on the extracted HRV features. Moreover, noise and short-term variability in the signals, which may confuse or degrade the performance of the classifiers, are better averaged out over longer observation periods. \Cref{tab:modelperf} demonstrates that a 120-seconds HRV observation window is already sufficient for the proposed method to showcase its capability. The confusion matrix of the model with the highest performance is shown in \cref{fig:CM-a}. A high differentiation is observed among the three classes, slightly better at detecting Moderate engagement than Low and High engagement. The F1 scores for Low, Moderate, and High levels of engagement are 0.92, 0.96, and 0.90, respectively. To improve performance in Low and High engagement, the model requires more balanced data or better features for class distinction. All in all, the \textit{DAiSEE} dataset's 10-seconds clips prevent an evaluation of the effects of HRV observation window length on engagement estimation. Yet, from the \textit{Engagement Dataset}, it's evident that HRV features can robustly measure engagement when given a sufficient observation window size, such as 2 to 4 minutes.

%-------------------------------------------------------------------------
\subsection{Effectiveness of BF selection when fusing with HRV for engagement estimation}

While HRV features have established a strong foundation, relying solely on physiological signals limits prediction accuracy. Adding BF is crucial for improved metrics as human engagement involves not only physiological but also subtle behavioral cues. Thus, a model fusing both physiological and behavioral aspects could enhance prediction.

In \cref{tab:big_comparison_table}, the ensemble of knn and Random Forest model's accuracy using both HRV and various BF combinations, including (1) gaze tracking, (2) angle of gaze direction, (3) 2D and 3D landmarks of specific points around each eye, (4) translation and rotation of the head in 3D space, and (5) facial Action Units, is shown. The ROC AUC metric, mirroring the accuracy trend, was omitted. The model's performance is significantly influenced by the BF selection. Set (3) is found to be less efficient with longer HRV observation windows; however, sets (4) and (5) are shown to boost prediction accuracy. The highest performance was observed when HRV was paired with (4), indicating "translation and rotation of the head in 3D space", and (5), pointing to "facial Action Units". Significant metric enhancement is observed when BFs are incorporated as the model's accuracy increases from 0.816 to 0.930 for a 60-seconds observation window. Enhanced performance due to BF inclusion is consistently observed across all data. Yet, the boost is more pronounced at smaller HRV observation window sizes. In \cref{fig:CM-b}, the confusion matrix of the knn+RF model, incorporating HRV and both (4) and (5) BF, is presented, as opposed to the HRV-only feature confusion matrix depicted in \cref{fig:CM-a}. It outperforms the singular HRV model, underscoring that the inclusion of BF yields superior engagement estimation. After fusing with BF, the F1 scores for Low, Moderate, and High levels of engagement improved to 0.93, 0.97, and 0.94, respectively. Further refinement through enhanced feature selection or the exploration of alternate models might offer improved differentiation between Low and High Engagement.

%-------------------------------------------------------------------------
\subsection{Combining HRV observation window size and BF set for engagement estimation}

\begin{figure}[b]
  \centering
   \includegraphics[width=0.8\linewidth]{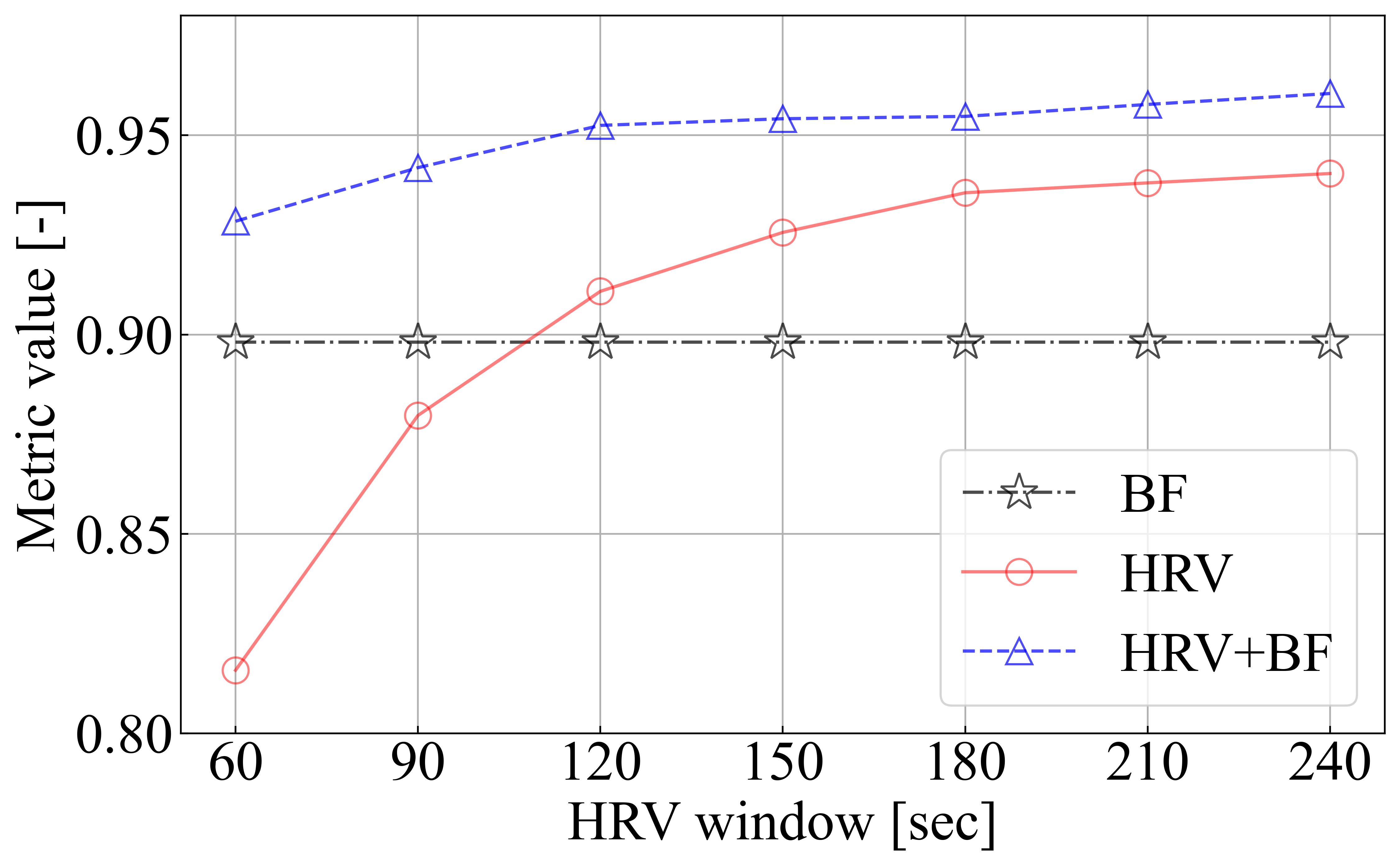}
   % {figures/HRV_BF_old.png}
   \caption{Accuracy of the ensemble of knn and Random Forest model for engagement estimation based on HRV and both (4) and (5) BF calculated at different values of HRV observation window. The accuracy of the model, based solely on BF, remains constant due to its independence from the value of the HRV window.}
   \label{fig:modelperf_behav}
\end{figure}

The effect of combining BF (4+5) set with different HRV observation window sizes on the model's performance is shown in \cref{fig:modelperf_behav}. As the HRV observation window size increases, the performance metrics of the model tends to improve, up to a certain point. However, the performance improvement isn't strictly linear and plateaus after a certain HRV observation window size. In the knn+RF model with BF (4+5), accuracy improves from 0.928 (60 seconds) to 0.960 (240 seconds), but the difference between 210 seconds and 240 seconds is marginal. Thus, while the integration of BF improves the model, gains are notably larger for smaller HRV observation windows. Nevertheless, with the addition of BF, the best HRV-only model's performance was enhanced by 2\%. \Cref{tab:starting_comparison} demonstrates the comparison of the proposed method with the method proposed by Mohamad \etal~\cite{mohamad2020automatic} which is publicly available and aimed for engagement estimation. The \textit{Engagement Dataset} was processed using their method, following the same protocol and data split as described in \cref{subsec:exp_protocol}. 

\begin{table}[t]
  \centering

  {\fontsize{9}{8.25}\selectfont
  \begin{tabular}{lcc}
    \toprule
    Method & Accuracy [-] & ROC AUC [-] \\
    \midrule
    Mohamad \etal~\cite{mohamad2020automatic} & 0.633 & 0.500 \\
    HRV & 0.940 & 0.983 \\
    HRV + BF (4+5) & \underline{\textbf{0.960}} & \underline{\textbf{0.991}} \\
    \bottomrule
  \end{tabular}
  \caption{Comparison of methods' performances for engagement estimation based on \textit{Engagement Dataset}. Ensemble of knn and Random Forest model was used for "HRV" and "HRV + BF (4+5)" methods.}
  \label{tab:starting_comparison}}
\end{table}
\section{Conclusion}

The remote measurement of heart rate variability, along with facial and body language recognition, is used for engagement analysis in virtual meetings, which have become increasingly popular over the recent years. This article pioneers the application of unsupervised rPPG measurement technology in estimating engagement during online meetings. It first introduces the \textit{Engagement Dataset} centered on social workers' online video meetings. The \textit{Engagement Dataset} is accompanied by granular engagement labels that capture the essence of virtual meeting dynamics. Subsequently, the effect of HRV observation window size on engagement estimation performance was assessed using both collected and public datasets. Short HRV observation windows proved to be unreliable for HRV feature calculation, while longer observation windows (e.g., 2-4 minutes) provided a robust foundation for engagement estimation. Further, the significance of selecting the right behavioral features set was evaluated. Given the crucial importance of the right HRV window choice, a performance increase from 49.40\% through 81.60\% to 94\% was witnessed when the HRV observation window size was adjusted from 10 seconds to 60 seconds and then to 240 seconds. Moreover, when the correct BF set was used, performance was further boosted by up to 2\%. 

In future, the proposed method is to be validated using more datasets. With the formulated method, engagement analysis at the clip level was carried out, setting the stage for future entire video recording level analysis. Additionally, exploration of engagement fluctuation on a group level, e.g., the analysis of synchrony or interactions of the participants, is planned to better analyze the online event. Moreover, this method is intended to be advanced into a deep learning model, trained end-to-end for engagement estimation.

\section*{Acknowledgments}
This work was supported by the Research Council of Finland (former Academy of Finland) Academy Professor project EmotionAI (grants 336116, 345122), and the Finnish Work Environment Fund (Project 200414). The authors also acknowledge CSC-IT Center for Science, Finland, for providing computational resources.

\newpage
{
    \small
    \bibliographystyle{ieeenat_fullname}
    \bibliography{main}
}
\clearpage
\setcounter{page}{1}
\maketitlesupplementary

\begin{figure*}[!ht]
  \centering
  \includegraphics[width=0.75\linewidth, height=0.15\textheight]{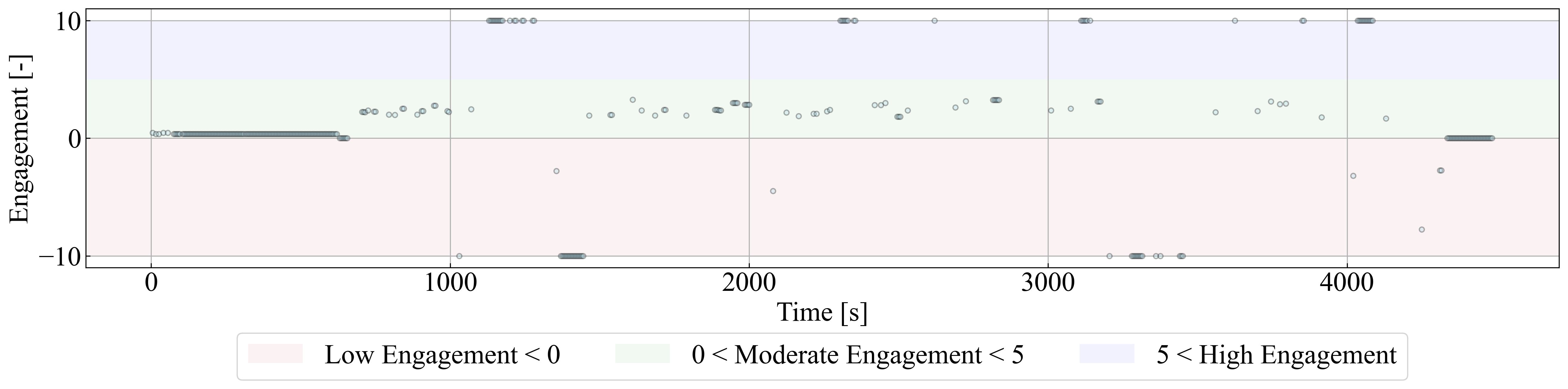}
  \caption{Example of engagement data obtained after filtering process. Data on this graph represent the engagement labels of a specific participant during a particular  video meeting.}
  \label{fig:eng_individual}
\end{figure*}

\begin{table}[!hb]
  \centering
    \caption{Description of Action Units used in engagement analysis.}
  {\fontsize{9}{10}\selectfont
  \begin{tabular}{llll}
    \toprule
    AU \# & Description & AU \# & Description \\
    \midrule
    AU1 & inner brow raiser & AU12 & lip corner puller \\
    AU2 & outer brow raiser & AU14 & dimpler \\
    AU4 & brow lowerer & AU15 & lip corner depressor \\
    AU5 & upper lid raiser & AU17 & chin raiser \\
    AU6 & cheek raiser & AU20 & lip stretcher \\
    AU7 & lid tightener & AU23 & lip tightener \\
    AU9 & nose wrinkler & AU25 & lips part \\
    AU10 & upper lip raiser & AU26 & jaw drop \\
    & & AU45 & blink \\
    \bottomrule
  \end{tabular}
  }
  \label{tab:AUDescription}
\end{table}

\end{document}